\documentclass[runningheads]{llncs}
\usepackage[T1]{fontenc}
\usepackage{graphicx}
\usepackage{booktabs}
\usepackage[misc]{ifsym}
\newcommand{\corr}{(\Letter)}
\usepackage{amsmath}
\usepackage{amssymb}
\usepackage{multirow}
\usepackage{hyperref}
\usepackage{microtype}
\usepackage{color}

\begin{document} 

\title{Beyond Synthetic Augmentation: Group‑Aware Threshold Calibration for Robust Balanced Accuracy in Imbalanced Learning}
\titlerunning{Group‑Aware Threshold Calibration}

\author{Hunter Gittlin\inst{1}\corr}

\institute{Pine Crest School, 1501 NE 62nd St, Fort Lauderdale, FL 33334, USA
\email{huntergittlin@gmail.com}}

\maketitle
\begin{abstract}

Class imbalance remains a fundamental challenge in machine learning, with traditional solutions often creating as many problems as they solve. We demonstrate that group-aware threshold calibration—setting different decision thresholds for different demographic groups—provides superior robustness compared to synthetic data generation methods. Through extensive experiments we show that group-specific thresholds achieve 1.5-4\% higher balanced accuracy than SMOTE and CT-GAN augmented models while improving worst-group balanced accuracy. Unlike single-threshold approaches that apply one cutoff across all groups, our group-aware method optimizes the Pareto frontier between balanced accuracy and worst-group balanced accuracy, enabling fine-grained control over group-level performance. Critically, we find that applying group thresholds to synthetically augmented data yields minimal additional benefit, suggesting these approaches are fundamentally redundant. Our results span seven model families including linear, tree-based, instance-based, and boosting methods, confirming that group-aware threshold calibration offers a simpler, more interpretable, and more effective solution to class imbalance.

\end{abstract}
\section{Introduction}

\begin{figure}[t]
  \centering
  \includegraphics[width=.49\linewidth]{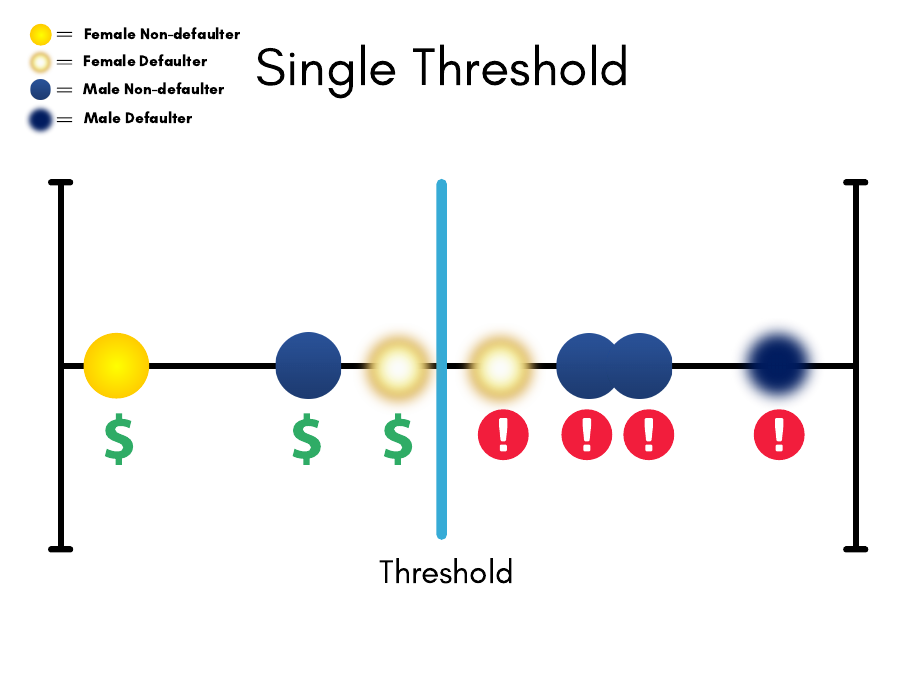}\hfill
  \includegraphics[width=.49\linewidth]{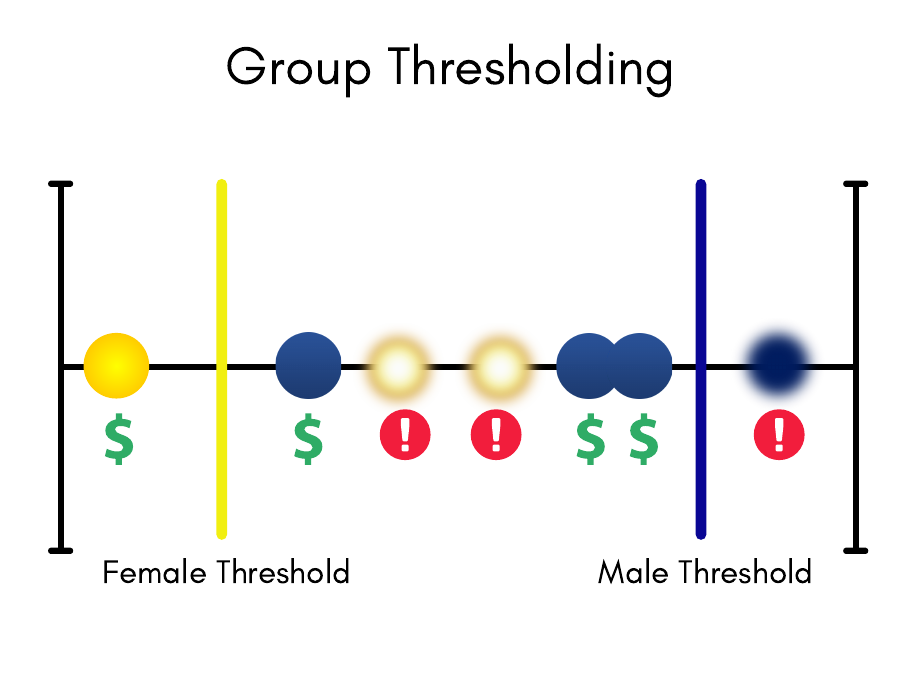}
  \caption{\textbf{Left:} Classification of male and female non-defaulters and defaulters in a single threshold setting.  This single threshold performs poorly, misclassifying three out of seven individuals (roughly 57\% accuracy).  \textbf{Right:} Classification of male and female non-defaulters and defaulters in a group thresholding scenario, in which there is a separate threshold for males and females.  In this case, the model correctly classifies all individuals as either defaulting or non-defaulting.}
  \label{fig:threshold_comparison}
\end{figure}

Algorithmic scores increasingly gate access to credit, welfare, and jobs. When they underrate certain groups they block housing, entrepreneurship, and civic participation, breaching fairness mandates such as the EU AI Act and the U.S. Equal Credit Opportunity Act. A frequent technical culprit is class imbalance: if 99\% of records are legitimate, a model that predicts no default everywhere enjoys 99\% accuracy while failing exactly where oversight matters. We propose a practical alternative—\emph{group-aware threshold calibration}—that assigns separate decision cut-offs to each protected group, outperforms heavyweight oversampling pipelines, and gives auditors an explicit lever over group-level error rates. Experiments on two financial benchmarks show that responsibly leveraging sensitive attributes in this way lifts both overall and worst-group balanced accuracy, reinforcing evidence that such attributes can be a powerful instrument for fairness. These results challenge blanket bans on their use, suggesting that tightly regulated access—rather than wholesale exclusion—may be essential for the equity goals those laws pursue, even as many practitioners still default to synthetic oversampling.

Synthetic approaches like SMOTE \cite{chawla_smote_2002} and CT-GAN \cite{xu_modeling_2019} attempt to balance datasets by creating artificial minority class samples. However, recent evidence suggests these methods introduce problematic artifacts. As demonstrated across 71 datasets, oversampling methods often lead to overfitting and poor generalization, with the authors concluding that oversampling should be avoided in real-world applications \cite{hassanat_stop_2022}. As these synthetic samples often create overlapping class regions that confuse decision boundaries rather than clarifying them.

We propose a fundamentally different approach: rather than manipulating training data and hoping for improved outcomes, we directly optimize for balanced accuracy through group-aware threshold calibration. Unlike single-threshold approaches that apply one decision boundary across all samples, group-aware thresholds recognize that different groups may require different decision criteria due to varying base rates or feature distributions \cite{hardt_equality_2016-1}. Fig.\@ \ref{fig:threshold_comparison} illustrates this concept concretely.

\section{Background and Related Work}

\subsection{The Limitations of Accuracy in Imbalanced Settings}

Traditional accuracy fails catastrophically with class imbalance.  Consider a dataset with 95\% negative and 5\% positive examples—a classifier predicting all negatives achieves 95\% accuracy while completely failing to identify any positive cases.  This paradox \cite{haibo_he_learning_2009} has motivated alternative metrics that give equal importance to all classes.

Balanced accuracy addresses this by averaging per-class accuracies:
\begin{equation}
BA = \frac{1}{2}\left(\frac{TP}{TP + FN} + \frac{TN}{TN + FP}\right) = \frac{TPR + TNR}{2}
\end{equation}

This formulation ensures that a trivial all-negative classifier achieves only 50\% balanced accuracy, properly reflecting its failure on the positive class.  Recent work by \cite{sagawa_distributionally_2020} connects balanced accuracy optimization to distributional robustness, providing theoretical foundations for its use in fairness-critical applications.  The connection between worst-group performance and distributional robustness has been further established by \cite{liu_just_2021}, who show that optimizing for worst-group accuracy provides guarantees against distribution shift.

\subsection{Synthetic Data Generation: Promise vs.  Reality}

SMOTE generates synthetic minority samples by interpolating between existing instances and their k-nearest neighbors.  While intuitive, this approach suffers from several fundamental limitations.  \cite{blagus_smote_2013} showed that SMOTE can increase classifier variance without improving minority class recognition in high-dimensional settings.  More recently, the comprehensive GHOST study by \cite{esposito_ghost_2021} tested 138 datasets and found that threshold optimization significantly outperformed SMOTE for 75\% of ML methods tested.

A critical limitation identified by \cite{kovacs_empirical_2019} is that SMOTE and similar oversampling methods lead to poorly calibrated probability estimates, with significantly worse Log-Loss and Brier scores compared to threshold-based approaches.  This calibration degradation is particularly problematic for applications requiring meaningful confidence scores, such as medical diagnosis or financial risk assessment.

CT-GAN attempts to address these issues through conditional generative adversarial networks, learning the underlying data distribution rather than simple interpolation.  However, \cite{engelmann_conditional_2020} found that even sophisticated generative models struggle with tabular data, often producing unrealistic feature combinations that don't respect complex dependencies in real-world datasets.  Our experiments confirm these findings, revealing that CT-GAN fails to improve upon simple threshold optimization despite its computational complexity.

\subsection{Group-Aware Threshold Optimization}

Traditional threshold optimization applies a single cutoff across all predictions.  However, when protected groups exhibit different base rates or feature distributions, this one-size-fits-all approach can perpetuate disparities.  Group-aware threshold optimization addresses this by learning separate thresholds for each demographic group, a approach that \cite{corbett-davies_algorithmic_2017,hardt_equality_2016-1} show can achieve optimal fairness-accuracy trade-offs under certain conditions.

Formally, for groups $g \in \{1, ..., G\}$ and predicted probabilities $p_i$, we learn thresholds $\tau_g$ such that:
\begin{equation}
\hat{y}_i = \mathbb{1}[p_i \geq \tau_{g(i)}]
\end{equation}

where $g(i)$ denotes the group membership of instance $i$.  This enables fine-grained control over group-level true positive and false positive rates.

\section{Methods}

\subsection{Experimental Setup}

We evaluate our approach on two benchmark datasets with natural class imbalance and protected groups.  The UCI Default of Credit Card Clients dataset contains 30,000 instances with 22.1\% default rate using sex as the protected attribute.  The Adult Income dataset includes 48,842 instances with 24.1\% high-income rate, using race as the protected attribute.

For each dataset, we implement four approaches.  The baseline uses models trained on original imbalanced data with a single threshold.  SMOTE applies group-aware oversampling to balance classes per demographic, implemented by separately generating synthetic samples for male and female subgroups to avoid information leakage between groups, following recommendations by \cite{delaney_oxonfair_2024}.  CT-GAN performs conditional generation with demographic conditioning, where we train the generative model for 5 epochs and condition on the protected attribute to generate synthetic positive samples proportional to each group's representation.  Finally, we apply group-aware threshold optimization to all data variants.

We employ 5-fold stratified cross-validation with 20\% test sets.  Within training data, we reserve 12.5\% for threshold optimization using OxonFair's grid search over group-specific thresholds, ensuring no data leakage \cite{delaney_oxonfair_2024}.  The validation set provides unbiased estimates for threshold selection. 

\subsection{Model Families}

To ensure findings generalize across algorithmic paradigms, we test several diverse classifiers spanning linear methods (Logistic Regression, SGD Classifier), tree-based approaches (Random Forest, Histogram-based Gradient Boosting, XGBoost, CatBoost), and instance-based methods (k-Nearest Neighbors).  This diversity ensures our conclusions aren't artifacts of specific algorithms.

\subsection{Evaluation Framework}

We focus on two key metrics that capture performance on imbalanced data with protected groups.  Balanced Accuracy (BA) serves as our primary metric for overall imbalanced performance, computed as the average of true positive rate and true negative rate to give equal weight to both classes.  As noted by \cite{luque_impact_2019}, balanced accuracy provides a more reliable assessment than traditional accuracy for imbalanced datasets, maintaining consistent interpretation across different class distributions.

Worst-Group Balanced Accuracy (WG-BA) measures the minimum balanced accuracy across demographic groups, ensuring no group is left behind by our optimization.  This metric aligns with the group distributional robustness framework of \cite{sagawa_distributionally_2020}, providing guarantees against performance degradation for minority groups.

For threshold optimization, we explore two objectives using group-specific thresholds.  Fair-BalAcc maximizes overall balanced accuracy while using different thresholds per group, allowing the algorithm to find the best global performance while leveraging group-specific decision boundaries.  Fair-MinBalAcc explicitly maximizes worst-group balanced accuracy, directly optimizing for the most disadvantaged group to ensure equitable performance.

\section{Results}

\subsection{Main Findings: Group-Aware Thresholds Dominate}

Tables ~\ref{tab:main_results} and \ref{tab:main_results_new} present comprehensive results across all model families, revealing that group-aware threshold optimization on original data consistently outperforms synthetic augmentation approaches. The pattern holds remarkably consistent across diverse algorithmic paradigms, confirming findings from \cite{hardt_equality_2016-1} that threshold optimization often provides more reliable improvements than data-level interventions.

\begin{table}[!htbp]
\centering
\caption{Results for UCI Default of Credit Card Clients dataset: Balanced accuracy (BA) and worst-group balanced accuracy (WG-BA) across all models and methods on credit default dataset. Bold indicates best overall performance within each model. \underline{Underlined} values show when Original+Thresholding outperforms both SMOTE-Raw and CTGAN-Raw baselines.}
\label{tab:main_results}
\scriptsize
\setlength{\tabcolsep}{3pt}
\begin{tabular}{@{}llcccccccc@{}}
\toprule
& & \multicolumn{2}{c}{Original Data} & \multicolumn{2}{c}{SMOTE} & \multicolumn{2}{c}{CT-GAN} \\
\cmidrule(lr){3-4} \cmidrule(lr){5-6} \cmidrule(lr){7-8}
Model & Method & BA & WG-BA & BA & WG-BA & BA & WG-BA \\
\midrule
\multirow{3}{*}{Logistic Reg.} 
 & Raw & 0.603 & 0.592 & 0.650 & 0.643 & 0.619 & 0.606 \\
 & Fair-BalAcc & \underline{\textbf{0.687}} & \underline{\textbf{0.683}} & 0.663 & 0.658 & 0.676 & 0.673 \\
 & Fair-MinBalAcc & \underline{0.686} & \underline{\textbf{0.683}} & 0.662 & 0.658 & 0.675 & 0.667 \\
\midrule
\multirow{3}{*}{SGD} 
 & Raw & 0.524 & 0.523 & 0.520 & 0.520 & 0.508 & 0.507 \\
 & Fair-BalAcc & \underline{\textbf{0.535}} & \textbf{0.522} & 0.526 & 0.511 & 0.518 & 0.505 \\
 & Fair-MinBalAcc & \underline{\textbf{0.535}} & \textbf{0.522} & 0.526 & 0.511 & 0.518 & 0.505 \\
\midrule
\multirow{3}{*}{Random Forest} 
 & Raw & 0.657 & 0.653 & 0.676 & 0.674 & 0.659 & 0.654 \\
 & Fair-BalAcc & \underline{0.700} & \underline{0.695} & 0.685 & 0.678 & 0.693 & 0.691 \\
 & Fair-MinBalAcc & \underline{\textbf{0.701}} & \underline{\textbf{0.695}} & 0.686 & 0.679 & 0.693 & 0.691 \\
\midrule
\multirow{3}{*}{Hist. GB} 
 & Raw & 0.657 & 0.655 & 0.674 & 0.672 & 0.656 & 0.654 \\
 & Fair-BalAcc & \underline{\textbf{0.709}} & \underline{0.704} & 0.676 & 0.672 & 0.706 & 0.699 \\
 & Fair-MinBalAcc & \underline{0.705} & \underline{\textbf{0.703}} & 0.675 & 0.672 & 0.703 & 0.699 \\
\midrule
\multirow{3}{*}{XGBoost} 
 & Raw & 0.648 & 0.643 & 0.654 & 0.652 & 0.650 & 0.647 \\
 & Fair-BalAcc & \underline{0.691} & \underline{0.684} & 0.664 & 0.660 & 0.690 & 0.684 \\
 & Fair-MinBalAcc & \underline{\textbf{0.692}} & \underline{\textbf{0.685}} & 0.663 & 0.660 & 0.690 & 0.683 \\
\midrule
\multirow{3}{*}{CatBoost} 
 & Raw & 0.656 & 0.654 & 0.669 & 0.665 & 0.656 & 0.653 \\
 & Fair-BalAcc & \underline{0.708} & \underline{0.705} & 0.674 & 0.670 & \textbf{0.709} & 0.705 \\
 & Fair-MinBalAcc & \underline{0.707} & \underline{0.701} & 0.674 & 0.671 & \textbf{0.710} & \textbf{0.706} \\
\midrule
\multirow{3}{*}{k-NN} 
 & Raw & 0.542 & 0.538 & 0.575 & 0.572 & 0.542 & 0.538 \\
 & Fair-BalAcc & 0.566 & 0.556 & \textbf{0.568} & 0.555 & 0.565 & 0.556 \\
 & Fair-MinBalAcc & 0.566 & 0.556 & \textbf{0.568} & 0.555 & 0.565 & 0.556 \\
\bottomrule
\end{tabular}
\end{table}

\begin{table}[!htbp]
\centering
\caption{Adult Income dataset: Balanced accuracy (BA) and worst-group balanced accuracy (WG-BA) across all models and methods on the credit-default dataset. \textbf{Bold} = best overall performance within each model. \underline{Underlined} = Original + Thresholding outperforms both SMOTE-Raw and CTGAN-Raw baselines.}
\label{tab:main_results_new}
\scriptsize
\setlength{\tabcolsep}{3pt}
\begin{tabular}{@{}llcccccccc@{}}
\toprule
& & \multicolumn{2}{c}{Original Data} & \multicolumn{2}{c}{SMOTE} & \multicolumn{2}{c}{CT-GAN} \\
\cmidrule(lr){3-4}\cmidrule(lr){5-6}\cmidrule(lr){7-8}
Model & Method & BA & WG-BA & BA & WG-BA & BA & WG-BA \\
\midrule
\multirow{3}{*}{Logistic Reg.} 
 & Raw              & 0.674 & 0.665 & 0.738 & 0.668 & 0.685 & 0.662 \\
 & Fair-BalAcc      & \underline{\textbf{0.753}} & \underline{\textbf{0.708}} & 0.747 & 0.674 & 0.750 & 0.691 \\
 & Fair-MinBalAcc   & \underline{\textbf{0.753}} & \underline{\textbf{0.708}} & 0.746 & 0.680 & 0.751 & 0.692 \\
\midrule
\multirow{3}{*}{SGD} 
 & Raw              & 0.558 & \textbf{0.550} & 0.557 & 0.550 & 0.541 & 0.534 \\
 & Fair-BalAcc      & \underline{\textbf{0.576}} & 0.500 & 0.575 & 0.500 & 0.543 & 0.517 \\
 & Fair-MinBalAcc   & \underline{\textbf{0.576}} & 0.500 & 0.575 & 0.500 & 0.543 & 0.517 \\
\midrule
\multirow{3}{*}{Random Forest} 
 & Raw              & 0.775 & 0.735 & 0.786 & 0.756 & 0.778 & 0.738 \\
 & Fair-BalAcc      & \underline{\textbf{0.815}} & 0.748 & 0.806 & 0.736 & 0.811 & 0.750 \\
 & Fair-MinBalAcc   & \underline{\textbf{0.815}} & 0.748 & 0.806 & \textbf{0.760} & 0.810 & 0.750 \\
\midrule
\multirow{3}{*}{Hist.\ GB} 
 & Raw              & 0.796 & 0.725 & 0.802 & 0.774 & 0.800 & 0.734 \\
 & Fair-BalAcc      & \underline{\textbf{0.836}} & 0.753 & 0.824 & 0.729 & 0.835 & 0.760 \\
 & Fair-MinBalAcc   & \underline{\textbf{0.836}} & 0.753 & 0.824 & 0.729 & 0.834 & 0.760 \\
\midrule
\multirow{3}{*}{XGBoost} 
 & Raw              & 0.797 & 0.741 & 0.805 & 0.757 & 0.799 & 0.754 \\
 & Fair-BalAcc      & \underline{\textbf{0.835}} & \underline{0.765} & 0.827 & 0.721 & 0.834 & 0.799 \\
 & Fair-MinBalAcc   & \underline{0.833} & \underline{0.765} & 0.827 & 0.721 & 0.834 & \textbf{0.800} \\
\midrule
\multirow{3}{*}{CatBoost} 
 & Raw              & 0.794 & 0.745 & 0.808 & 0.760 & 0.796 & 0.732 \\
 & Fair-BalAcc      & \underline{\textbf{0.838}} & \underline{0.761} & 0.830 & 0.757 & 0.838 & 0.761 \\
 & Fair-MinBalAcc   & \underline{\textbf{0.838}} & \underline{\textbf{0.779}} & 0.828 & 0.757 & 0.838 & 0.769 \\
\midrule
\multirow{3}{*}{k-NN} 
 & Raw              & 0.611 & 0.594 & 0.608 & 0.561 & 0.612 & 0.587 \\
 & Fair-BalAcc      & \underline{0.614} & \underline{\textbf{0.596}} & \textbf{0.618} & 0.586 & 0.612 & 0.553 \\
 & Fair-MinBalAcc   & \underline{0.614} & \underline{\textbf{0.596}} & 0.614 & 0.559 & 0.613 & 0.557 \\
\bottomrule
\end{tabular}
\end{table}

On the \textbf{UCI Default of Credit Card Clients dataset}, for logistic regression, group-aware threshold optimization on original data achieves a balanced accuracy of $0.687$, outperforming both SMOTE's raw performance ($0.650$) and CT-GAN's raw performance ($0.619$). Applying fairness thresholds to the synthetic data from SMOTE and CT-GAN results in balanced accuracies of $0.663$ and $0.676$ respectively, both of which fall short of the results from applying thresholds to the original data. The worst-group balanced accuracy tells a similar story, with original data plus thresholds achieving $0.683$, substantially exceeding SMOTE-Raw ($0.643$) and CTGAN-Raw ($0.606$).

Tree-based models on the same dataset demonstrate even stronger patterns. For Hist.\ GB, group-aware thresholds on original data achieve a balanced accuracy of $0.709$, a significant improvement over SMOTE's raw performance ($0.674$) and CT-GAN's raw performance ($0.656$). The worst-group balanced accuracy for Hist.\ GB on original data with thresholds reaches $0.703-0.704$, which is notably better than SMOTE's raw performance ($0.672$).

\subsection{The Redundancy of Synthetic Augmentation}

A critical insight emerges when examining the incremental benefit of applying group-aware thresholds to synthetically augmented data. On the \textbf{Adult Income dataset}, for the CatBoost model, applying thresholds to the original data provides a large boost in balanced accuracy from $0.794$ to $0.838$ (a gain of $0.044$). However, when applied to the already-augmented SMOTE data, thresholds only provide a gain of $0.022$ (from $0.808$ to $0.830$). This pattern of diminishing returns holds across model families, supporting \cite{santos_cross-validation_2018} observation that combining multiple imbalance-handling techniques often yields limited additional benefits.

Similarly, for XGBoost on the credit default data, thresholding the original data increases balanced accuracy by $0.044$ (from $0.648$ to $0.692$). The same process on SMOTE data yields a meager gain of just $0.009$ (from $0.654$ to $0.663$). While CT-GAN augmented data sometimes shows larger gains from thresholding (e.g., for logistic regression on the credit data, BA improves from $0.619$ to $0.676$), this is often because its base performance is poor, and the final result still underperforms simple threshold optimization on the original data ($0.676$ vs. $0.687$).

The implication aligns with theoretical analysis by \cite{wallace_class_2011}, who showed that sampling and threshold-moving address the same underlying optimization problem through different mechanisms. Our empirical results confirm their theoretical prediction that these approaches prove largely redundant when combined.

\section{Discussion}

\subsection{Why Group-Aware Thresholds Succeed}

Rather than hoping synthetic data indirectly improves balanced accuracy across groups, threshold methods directly optimize the target metric.  This alignment proves more effective than proxy approaches that assume balancing training data will automatically improve group-specific and class-specific performance. Further, avoiding distribution shift maintains the integrity of the original data distribution.  While synthetic augmentation fundamentally alters training distributions with the intention of improving representation, this shift can degrade calibration and introduce artifacts that harm generalization, particularly for groups with different feature distributions.  .

\subsection{Implications for Practice}

Our findings suggest a revised workflow for handling imbalanced datasets with protected groups.  Practitioners should start with group-aware threshold optimization on original data, as it provides immediate improvements with minimal computational cost.  Comprehensive evaluation using balanced accuracy and worst-group metrics, not just overall accuracy, reveals the true performance across different populations.  Synthetic methods should be considered only when threshold optimization proves insufficient, such as in cases of extreme imbalance.

If synthetic augmentation is used, group-aware thresholds should still be applied, though our results suggest expecting minimal additional gains.  This approach prioritizes interpretability and efficiency while achieving superior performance.  Stakeholders can understand different confidence requirements for different groups, while avoiding the black-box nature of synthetic data generation.  

\subsection{Limitations and Future Work}

Several limitations warrant discussion.  Our experiments focus on binary classification with binary protected attributes, and multi-class imbalance or continuous protected attributes may show different patterns.  The datasets examined have moderate imbalance ratios (approximately 4:1), and extreme imbalance might benefit more from synthetic approaches.  Domain-specific constraints, such as regulatory requirements for equal treatment, may mandate certain approaches regardless of empirical performance.

Future work should explore theoretical analysis of when synthetic methods might outperform threshold optimization, perhaps in extreme imbalance scenarios or with specific data characteristics.  Extension to multi-class and multi-label settings would broaden applicability, as would handling multiple intersecting protected attributes.  

\section{Conclusion}

Class imbalance remains a pervasive challenge in machine learning, particularly when combined with fairness constraints across protected groups.  Our work demonstrates that group-aware threshold calibration provides a simple, interpretable, and effective solution that can outperform complex synthetic data generation approaches.  By setting different decision thresholds for different demographic groups, we achieve superior balanced accuracy and worst-group performance compared to SMOTE and CT-GAN augmentation.

The key insight is that synthetic augmentation and threshold optimization are fundamentally redundant—both attempt to address class imbalance, but threshold methods do so more directly and effectively.  This finding has important implications for the field, suggesting that much of the complexity introduced by synthetic data generation may be unnecessary.  For practitioners, the message is clear: before investing computational resources in synthetic data generation, explore group-aware threshold calibration.  Not only does it achieve better performance with orders of magnitude less computation, but it also provides transparent, interpretable fairness mechanisms that stakeholders can understand and trust.

\bibliographystyle{splncs04} 
\bibliography{bib}

\end{document}